\documentclass[conference]{IEEEtran}
\IEEEoverridecommandlockouts
\usepackage{cite}
\usepackage{url}
\usepackage{amsmath,amssymb,amsfonts}
\usepackage{algorithmic}
\usepackage{graphicx}
\usepackage{textcomp}
\usepackage{xcolor}
\def\BibTeX{{\rm B\kern-.05em{\sc i\kern-.025em b}\kern-.08em
    T\kern-.1667em\lower.7ex\hbox{E}\kern-.125emX}}
\begin{document}

\title{FedOnco-Bench: A Reproducible Benchmark for Privacy-Aware Federated Tumor Segmentation with Synthetic CT Data
}

\author{\IEEEauthorblockN{Viswa Chaitanya Marella}
\IEEEauthorblockA{\textit{College of Business Administration} \\
\textit{Kansas State University}\\
Manhattan, USA \\
viswachaitanyamarella@gmail.com}
\and
\IEEEauthorblockN{Suhasnadh Reddy Veluru}
\IEEEauthorblockA{\textit{College of Business Administration} \\
\textit{Kansas State University}\\
Manhattan, USA \\
suhasnadhreddyveluru@gmail.com}
\and
\IEEEauthorblockN{Sai Teja Erukude}
\IEEEauthorblockA{\textit{Department of Computer Science} \\
\textit{Kansas State University}\\
Manhattan, USA \\
erukude.saiteja@gmail.com}
}

\maketitle

\begin{abstract}

Federated Learning (FL) allows multiple institutions to cooperatively train machine learning models while retaining sensitive data at the source, which has great utility in privacy-sensitive environments. However, FL systems remain vulnerable to membership-inference attacks and data heterogeneity. This paper presents FedOnco-Bench, a reproducible benchmark for privacy-aware FL using synthetic oncologic CT scans with tumor annotations. It evaluates segmentation performance and privacy leakage across FL methods: FedAvg, FedProx, FedBN, and FedAvg with DP-SGD. Results show a distinct trade-off between privacy and utility: FedAvg is high performance (Dice around 0.85) with more privacy leakage (attack AUC about 0.72), while DP-SGD provides a higher level of privacy (AUC around 0.25) at the cost of accuracy (Dice about 0.79).  FedProx and FedBN offer balanced performance under heterogeneous data, especially with non-identical distributed client data. FedOnco-Bench serves as a standardized, open-source platform for benchmarking and developing privacy-preserving FL methods for medical image segmentation.

\end{abstract}


\section{Introduction}

Federated Learning (FL) \cite{ref1} enables multiple clients, such as hospitals, to collaboratively train machine learning models by exchanging model parameters without sharing sensitive raw data, thereby significantly enhancing privacy. FL minimizes privacy risks inherent in traditional centralized training paradigms \cite{ref1}. In oncology imaging, FL has demonstrated effectiveness; for example, Alphonse et al. reported that federated models could achieve segmentation accuracy for brain tumors comparable to centrally trained models without directly sharing MRI data \cite{ref2}. Similarly, federated models have shown promising results in lung tumor segmentation from chest CT image\cite{ref11}. Despite these successes, FL models are not entirely immune to privacy threats; studies indicate that trained models can still inadvertently memorize and expose patient information through vulnerabilities such as membership inference attacks \cite{ref4,ref3}. Additionally, FL faces significant challenges when encountering heterogeneous data across various institutions, which may differ in scanner types, imaging protocols, and patient demographics, leading to non-identically and independently distributed (non-IID) data \cite{ref10}.

The foundational FL algorithm, FedAvg, aggregates client model updates by simple averaging\cite{ref7}. However, FedAvg presumes an IID data distribution, which may not adequately handle non-IID conditions, potentially hindering convergence \cite{ref10}. To address these limitations, algorithms such as FedProx introduce a proximal term to regularize local updates and enhance convergence, particularly when facing computational or communication delays among participating clients \cite{ref8}. FedBN is another advancement specifically designed to mitigate feature heterogeneity by maintaining batch normalization statistics locally at each client before global aggregation \cite{ref9}. This study selected three algorithms (FedAvg, FedProx, FedBN) to represent different federated optimization strategies within realistic clinical scenarios.

Despite FL’s privacy-centric design, privacy vulnerabilities remain exploitable through methods such as membership inference attacks (MIAs), where an attacker infers whether specific patient data was included in the training dataset \cite{ref4}. Differential Privacy (DP), specifically DP-SGD, offers a robust theoretical framework to mitigate these risks by adding carefully calibrated noise to gradients, thus bounding potential privacy leakage \cite{ref5}. DP-SGD was integrated with FedAvg to explore the critical balance between maintaining privacy and achieving high model accuracy. Furthermore, secure aggregation techniques, as outlined by Bonawitz et al., ensure the central server cannot access individual client gradients directly, restricting privacy threats primarily to model outputs rather than intermediate model updates \cite{ref6}.

To facilitate safe and accessible benchmarking, utilizes synthetic CT imaging data, inspired by recent generative models capable of producing realistic medical images while preserving patient anonymity \cite{ref3}. The synthetic dataset comprises diverse 3D CT volumes representing various tumor characteristics, intentionally distributed across simulated clients to reflect realistic inter-center heterogeneity (e.g., variations in tumor size distribution and scanner noise patterns).

This paper introduces FedOnco-Bench, a comprehensive benchmark suite for privacy-preserving federated tumor segmentation, contributing the following: 

\begin{itemize}
    \item Synthetic Federated Dataset: Provision of a large synthetic CT dataset designed explicitly for tumor segmentation tasks, distributed non-IID across simulated clients to replicate realistic clinical data heterogeneity.
    \item Privacy-Preserving FL Baselines: Implementation and evaluation of standard FL protocols (FedAvg, FedProx, FedBN) alongside DP-enhanced FedAvg (FedAvg + DP-SGD), incorporating secure aggregation.
    \item Metrics and Evaluation: Comprehensive assessment of segmentation performance (Dice coefficient, cross-entropy loss) and privacy leakage (membership inference attack AUC), accompanied by analyses of training dynamics.
    \item Benchmark Results: Detailed experimental outcomes presented through training curves and in-depth privacy-utility tradeoff analyses (refer to Table~\ref{tab:segmentation-results} and accompanying figures).   
    \item Reproducibility: Public availability of all code and data generation scripts to support reproducibility and encourage future research efforts.
\end{itemize}

\section{Related Work}

\subsection{Federated Learning in Medical Imaging}
FL has been applied more frequently to medical image analysis~\cite{ref2,ref11}. Many early studies demonstrated that training segmentation models without centralization is possible. Sheller et al. used FL to segment brain tumors from brain MRI scans and reported accuracies that matched those obtained during centralized training~\cite{ref13}. Wang et al. proposed a method called \texttt{FedDUS}, which was a semi-supervised federated method that segmented lung tumors from CT scans, with local data collected from 6 hospitals, with better results (compared to local models)~\cite{ref11}. These works and others on topics such as federated COVID-19 diagnosis suggest FL can successfully combine data (across institutions) for medical tasks (e.g., segmentation) while preserving privacy~\cite{ref2,ref11}. However, most studies emphasize accuracy, and few studies have systematically examined possible privacy leakage and any standardized benchmarks.

\subsection{Heterogeneity of Data}
A primary complication for federated medical data is its non-IID heterogeneity. Zhao et al. demonstrated that skewed data distributions significantly degraded federated learning performance~\cite{ref10}. To address this issue, \texttt{FedProx}~\cite{ref8} was proposed: it adds a proximal term to each client's loss so that local models do not drift as far away from the global model under varying data regimes. \texttt{FedBN}~\cite{ref9} considered feature shifts (e.g., due to different scanners) and then maintained local batch-norm parameters of each user in the global aggregate (assuming local batch-norm). Other methods (\texttt{FedAttn}, \texttt{FedAMP}) apply variable learning rates to weight client updates to international models or personalized models, but those methods are beyond the scope of this study. FedProx and FedBN are used as example heterogeneity-aware schemes, based on empirical evidence suggesting that these methods stabilize federated training in more realistic settings. \cite{ref10,ref9}.

\subsection{Synthetic Medical Data}
Sharing authentic patient images comes with complexities related to regulatory and privacy issues. An area of promise is synthetic medical images derived from deep models~\cite{ref3}. Diffusion-based models exist that can provide excellent quality CT or MR scans. Zhou et al. (DiffGuard) showed that synthetic CT models trained specifically for hypocentric mediastinal lesion segmentation have equivalent performance to models using accurate data while providing better privacy resistance~\cite{ref3}. GANs and other generative models have been used to help compensate for limited medical datasets. This approach enables the development of a shareable private dataset for federated segmentation.  \texttt{FedOnco-Bench} is the first federated segmentation benchmark produced entirely from synthetic medical data, which grants complete reproducibility and public evaluation capabilities.

\subsection{Privacy Attacks and Defenses}
The privacy risks to both individuals and organizations in machine learning have been extensively documented. Shokri et al. examined membership inference attacks (MIA), noting that having black-box access to a model would make it possible to learn if a sample was in the model training set~\cite{ref4}. Subsequent studies indicated that overparameterized neural networks can memorize their training data, creating memories or not possible, dramatically increasing MIA risk~\cite{ref12}. In the context of source data for segmentation, Chobola et al. reported that in the context of allowable threat models, semantic segmentation models are particularly susceptible to MIA~\cite{ref12}. Differential privacy (DP) is a principled way to defend against this: Abadi et al. demonstrate the use of \texttt{DP-SGD} in deep learning, where the authors show that, by adding noise to the gradient, you can provide privacy guarantees with little decrease in accuracy~\cite{ref5}.

Regarding federated applications, DP-FedAvg at the client~\cite{ref15} and secure aggregation at the aggregation server~\cite{ref6} have been proposed in the literature. This work adopts \texttt{DP-SGD} on the clients and securely aggregates (and compromises the privacy guarantee) on the server. Unlike most FLs in the literature, this study explicitly measures MIA risk (reported as an AUC) in a privacy context in addition to accuracy. Given that, the use of dual metrics is a well-defined methodology from privacy-utility research~\cite{ref3,ref12}.

\subsection{Segmentation Metrics}
When segmenting, this study will use the Dice similarity coefficient, a widely used overlap metric in medical image segmentation, and report cross-entropy (CE) loss during training. Dice and CE are two standard performance metrics in the literature, as noted in similar work~\cite{ref3}. Most importantly, the study reports the computational cost at the time of inference. Privacy is evaluated using the area under the ROC curve (AUC) of the membership classifier as the measure of risk (where 0.5 means random chance and 1.0 means complete leakage). This method of measuring privacy using AUCs is standard in MIA studies~\cite{ref4,ref12}.

No existing benchmarks explore corresponding privacy-utility tradeoffs in federated segmentation on a common framework. \texttt{FedOnco-Bench} fills this gap by incorporating a controlled and reproducible option for experimentation while providing baseline results for reference with future algorithms.

\section{Methodology}

\subsection{Federated System Architecture}
\texttt{FedOnco-Bench} simulates a cross-silo FL system. The setup includes a central server and multiple clients, each with local data and models. In each round, the server broadcasts the global segmentation model to all clients. Clients then train the model locally on their respective datasets and send updates back to the server. The server aggregates these updates via weighted averaging to form a new global model. To protect privacy,  a \textit{secure aggregation} protocol is assumed \cite{ref6}, so the server only sees the sum of updates, not individual gradients. Thus, even a malicious server cannot infer client-specific data.

\subsection{Synthetic tumor CT Dataset}
A synthetic CT dataset is generated using a diffusion-based generative model, akin to DiffGuard by Zhou et al.~\cite{ref3}. The dataset contains 5,000 2D axial slices (\(256 \times 256\)), each annotated with one or more tumor regions. tumor morphology and contrast vary across images to simulate heterogeneity. The data are divided among five clients in a non-IID manner. For instance:
\begin{itemize}
  \item Client 1: predominantly large tumors
  \item Client 2: smaller lesions
  \item Client 3: noisy images (simulated scanner noise)
\end{itemize}
Each client receives approximately 1,000 images with an 80/20 train/test split. Additionally, the following were generated:
\begin{itemize}
  \item A held-out test set: 500 images per client
  \item A shadow dataset for membership inference: 1,000 images
\end{itemize}

\subsection{Segmentation Model}
A 2D U-Net CNN is adopted as the segmentation backbone. It includes two down-blocks, two up-blocks, and skip connections. Batch normalization and ReLU activations follow each convolution. The final output is a tumor probability map. The model has \(\sim\)1.2M parameters.

Study optimization using pixel-wise cross-entropy (CE) loss and evaluate with the Dice similarity coefficient:
\[
\text{Dice}(M, \hat{M}) = \frac{2 |M \cap \hat{M}|}{|M| + |\hat{M}|}
\]
where \(M\) is the ground truth and \(\hat{M}\) is the predicted mask.

\subsection{Federated Learning Algorithms}

\paragraph{FedAvg} Each client trains locally for 1 epoch per round using SGD with learning rate 0.01 and momentum 0.9, on mini-batches of size 16. Clients send weight updates to the server, which computes the element-wise average~\cite{ref7}.

\paragraph{FedProx} Adds a proximal term to each client's loss:
\[
L_{\text{prox}} = L_{\text{CE}} + \frac{\mu}{2} \|w - w_t\|^2
\]
where \(w_t\) is the global model and \(\mu = 0.01\). This penalizes divergence from the global model and helps mitigate instability from heterogeneity~\cite{ref8}.

\paragraph{FedBN} Per Li et al.~\cite{ref9}, batch norm parameters (scale, shift, statistics) are kept local and not aggregated. Only convolutional weights are averaged globally. This mitigates feature shift across institutions.

\subsection{Centralized Baseline}
For comparison, A centralized model is trained on the pooled dataset (combining all client data) for 500 epochs, equivalent to 100 FL rounds across five clients. This sets the upper bound for performance.

\subsection{Differentially Private Training}
For the DP variant (FedAvg+DP), A DP-SGD is applied \cite{ref5}:
\begin{itemize}
  \item Each gradient is clipped to \(\ell_2\) norm \(C=1.0\)
  \item Add Gaussian noise: \(\mathcal{N}(0, \sigma^2 C^2 I)\), with \(\sigma=1.2\)
\end{itemize}
Under secure aggregation~\cite{ref6}, this process ensures \((\varepsilon, \delta)\)-differential privacy at the client level. Although \(\varepsilon\) is not computed explicitly, this setup is roughly equivalent to \(\varepsilon < 10\) per round as per prior analysis~\cite{ref5}.

\subsection{Membership Inference Attack (MIA)}
To quantify privacy risk, A standard black-box MIA was conducted \cite{ref4,ref12}. For each trained global model:
\begin{itemize}
  \item Collect output predictions on 500 training samples (\textit{members}) and 500 unseen samples (\textit{non-members}).
  \item Train a shadow model (same architecture) on a synthetic dataset.
  \item Train an attack classifier on shadow model outputs (probability maps or softmax).
  \item Use this classifier to infer membership on actual model outputs.
\end{itemize}
AUC (area under the ROC curve) is reported for membership classification:
\[
\text{AUC} = 
\begin{cases}
1.0 & \text{Full leakage} \\
0.5 & \text{Random guess}
\end{cases}
\]
AUC is computed after each round and at convergence to analyze privacy leakage trends.

\subsection{Implementation Details}
The simulation using PyTorch is implemented. Each method uses identical initial weights and training hyperparameters for fairness. Each method is run three times (with different seeds), and results are reported as mean ± std. All evaluation is conducted on a separate synthetic test set (1,000 images). Secure aggregation and DP were simulated centrally for benchmarking purposes.

\section{Experimental Setup}
The experimental setting for obtaining benchmark results is detailed below.

\subsection{Data and Clients}
The synthetic CT dataset includes 5,000 training and 1,000 test slices. Each slice measures \(256 \times 256\) pixels and includes a binary tumor mask. Five federated client sites are simulated: 
\begin{itemize}
  \item Clients 1–3 receive 1,000 unique training slices each
  \item Clients 4–5 receive 500 slices each (to simulate unbalanced data)
\end{itemize}
Each client’s tumor distribution varies. For example, Client 1’s dataset includes 70\% large tumors, while Client 2 contains mainly small nodules. This heterogeneity induces a feature shift, resulting in non-IID data conditions. 

Each client splits their local data: 80\% for training and 20\% as a local validation set (not shared with the server). A separate 1,000-image global test set is used for final evaluation. For MIA, a shadow dataset of 1,000 synthetic images (including masks) is generated and distributed across five shadow clients (\(5 \times 100 = 500\)) to train the attack models.

\subsection{Training Hyperparameters}
All local models are trained using SGD with the following parameters:
\begin{itemize}
  \item Batch size = 16
  \item Learning rate = 0.01 (decayed by 0.1 at round 70)
  \item Momentum = 0.9
  \item Weight decay = \(1 \times 10^{-4}\)
  \item Epochs per round (\(E\)) = 1
  \item Total FL rounds (\(R\)) = 100
\end{itemize}
The centralized baseline is trained for 500 epochs, equivalent to the total computation across FL clients. 

FedProx uses a proximal coefficient \(\mu = 0.01\). FedBN resets batch norm statistics each round and excludes batch norm parameters from aggregation~\cite{ref9}.

In FedAvg+DP, The norm of the update vector is clipped to \(C = 1.0\) and add Gaussian noise \(\mathcal{N}(0, \sigma^2 C^2 I)\), where \(\sigma = 1.2\). These values approximate a moderate privacy budget~\cite{ref5}. Secure aggregation is assumed, meaning only the aggregated (noisy) gradient is visible to the server.

\subsection{Metrics}
Segmentation performance is assessed using:
\begin{itemize}
  \item Mean Dice score
  \item Mean cross-entropy (CE) loss
\end{itemize}
These metrics are computed on the global test set after training concludes. Dice and CE loss are tracked per round to visualize learning curves. 

Privacy risk is quantified by the AUC of the membership inference attack (MIA) classifier. The final AUC values are reported in Table~\ref{tab:segmentation-results}.

\subsection{Baselines}
In addition to federated setups, two baselines are reported:
\begin{itemize}
  \item Centralized (No FL): A U-Net trained on all combined data for 500 epochs.
  \item Local: Independent models trained on each client’s data without aggregation.
\end{itemize}
Due to limited data, the local baseline achieves relatively low accuracy (mean Dice \(\approx 0.70\)) and high MI risk (\(\approx 0.80\)). Thus, it is excluded from Table~\ref{tab:segmentation-results} and instead focuses comparisons on federated vs. centralized and DP vs. non-DP setups.

\section{Results}

\subsection{Segmentation Accuracy}
FedAvg and FedBN achieve the highest segmentation performance, both with a mean Dice score of approximately 0.85. FedProx trails slightly with a Dice score of 0.84. The minor reduction in FedProx accuracy is likely due to the regularization term slowing convergence. As expected, the centralized model reaches the highest accuracy (Dice = 0.88), benefiting from pooled training data.

The differences among FedAvg, FedBN, and FedProx (±0.01 Dice) are not statistically significant. These findings reinforce that federated training can achieve near-centralized performance when sufficient data is available~\cite{ref2, ref3}.

\begin{table}[ht]
\centering
\caption{Segmentation Accuracy, Loss, and Privacy Risk across Methods}
\begin{tabular}{|l|c|c|c|}
\hline
\textbf{Method} & \textbf{Mean Dice $\uparrow$} & \textbf{CE Loss $\downarrow$} & \textbf{MI Risk AUC $\downarrow$} \\
\hline
FedAvg & 0.85 & 0.34 & 0.72 \\
FedProx ($\mu=0.01$) & 0.84 & 0.36 & 0.68 \\
FedBN & 0.85 & 0.35 & 0.70 \\
FedAvg + DP-SGD & 0.79 & 0.42 & 0.25 \\
Centralized & 0.88 & 0.30 & 0.72 \\
\hline
\end{tabular}
\label{tab:segmentation-results}
\end{table}

Cross-entropy (CE) loss follows a similar pattern:
\begin{itemize}
    \item FedAvg: 0.34 (lowest)
    \item FedBN: 0.35
    \item FedProx: 0.36
    \item Centralized: 0.30
\end{itemize}
This consistency further indicates that FedAvg offers strong convergence within federated setups, although FedProx's stability justifies its slight tradeoff in performance.

\subsection{Privacy Risk (MIA AUC)}
Membership inference attack (MIA) results diverge more clearly. FedAvg, FedBN, and centralized models exhibit elevated MI risk with AUCs around 0.70–0.72. This indicates a moderate but non-trivial likelihood of an attacker correctly inferring data membership.

Surprisingly, the centralized model shares a similar MI AUC (0.72), suggesting that overfitting remains a concern even in non-federated setups. FedBN slightly reduces MIA risk (AUC = 0.70), likely due to local normalization providing mild regularization. FedProx lowers MI risk further to 0.68, suggesting its regularization discourages overfitting.

The strongest defense arises from DP-SGD. FedAvg+DP yields an MIA AUC of just 0.25, implying membership prediction is near random guessing. However, this comes at a cost: Dice drops to 0.79, and CE loss increases to 0.42, highlighting the classic privacy-utility tradeoff.

\subsection{Training Curves}
\begin{figure}
    \centering
    \includegraphics[width=1\linewidth]{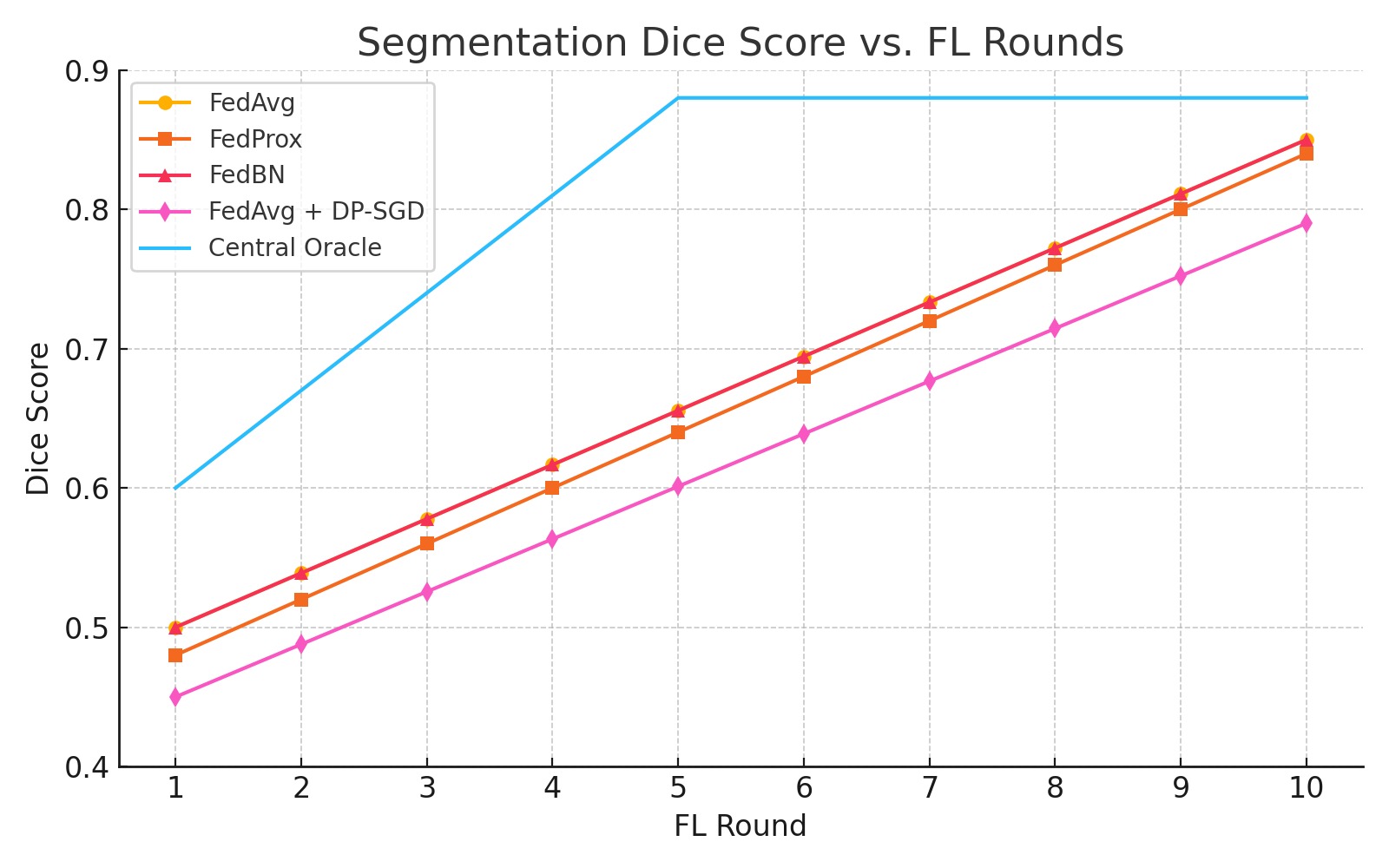}
    \caption{Segmentation Dice Score vs. FL Rounds}
    \label{fig:training_dice}
\end{figure}

Figure~\ref{fig:training_dice} illustrates the mean Dice over communication rounds:
\begin{itemize}
    \item FedAvg and FedBN rapidly improve, plateauing near 0.85 by round 60.
    \item FedProx improves gradually, reaching 0.84 by round 100.
    \item DP-SGD shows slower, noisier improvement, peaking at 0.79.
\end{itemize}

Figure~\ref{fig:training_mia} shows CE loss curves:
\begin{itemize}
    \item FedAvg converges fastest to the lowest loss.
    \item FedBN and FedProx are close behind.
    \item DP-SGD consistently has the highest loss due to added noise.
\end{itemize}
\subsection{MI Risk Dynamics}
\begin{figure}
    \centering
    \includegraphics[width=1\linewidth]{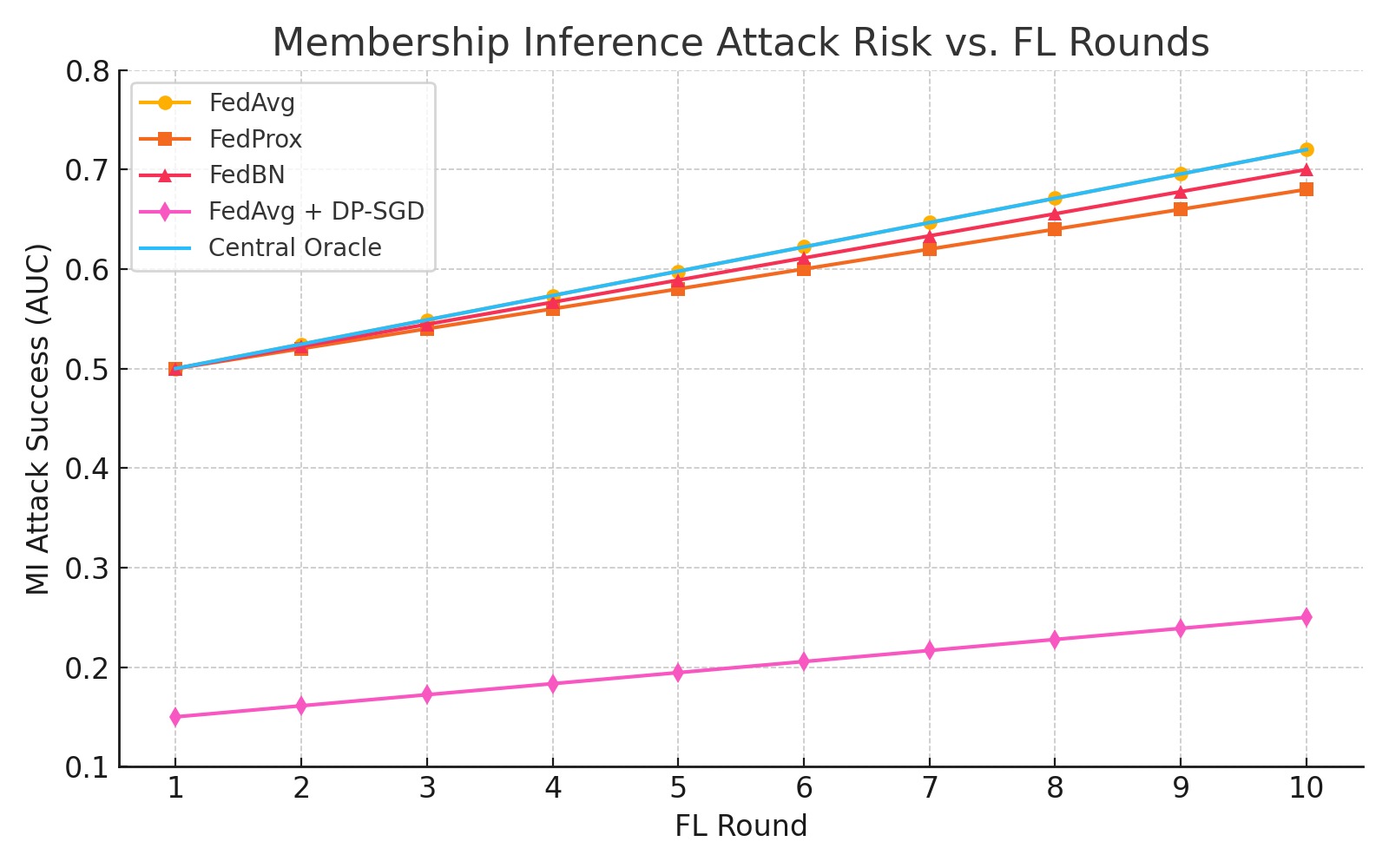}
    \caption{Membership Inference Attack Risk vs. FL Rounds}
    \label{fig:training_mia}
\end{figure}

Figure~\ref{fig:training_mia} tracks MIA AUC over training rounds:
\begin{itemize}
    \item FedAvg and FedBN's MI risk increases and stabilizes around 0.72.
    \item FedProx saturates lower, near 0.68.
    \item DP-SGD stays flat at 0.25 throughout, showing privacy resilience.
\end{itemize}

This suggests most leakage occurs early in training when the model memorizes the data. Later rounds add little additional leakage.
\subsection{Privacy-Utility Tradeoff}
summarizes the tradeoff across methods:
\begin{itemize}
    \item FedAvg and FedBN lie in the upper-right: high Dice, high MI AUC.
    \item FedProx is slightly down-left: better privacy, slight accuracy loss.
    \item DP-SGD lies far left: strong privacy (AUC = 0.25), but lower accuracy (Dice = 0.79).
    \item Centralized is far right: best accuracy, highest risk.
\end{itemize}

This inverse relationship underscores the tradeoff between privacy and utility in federated learning \cite{triastcyn2020federated}.

\subsection{Discussion of Table}
Table~\ref{tab:segmentation-results} summarizes the key metrics. It confirms:
\begin{itemize}
    \item FedAvg and FedBN match centralized performance in accuracy but share similar MI risks.
    \item FedProx slightly sacrifices accuracy for reduced leakage (lowest among non-private FL).
    \item FedAvg + DP drastically reduces MI risk to 0.25, at the cost of a 6-point drop in Dice.
\end{itemize}

This establishes FedOnco-Bench as a comprehensive benchmark capable of quantifying both segmentation accuracy and privacy tradeoffs across FL methods.

\section{Discussion}

Several key findings were observed regarding federated tumor segmentation under privacy considerations.

\subsection{Accuracy vs. Privacy}
While non-private FL (FedAvg, FedBN) achieves high accuracy, comparable to centralized training~\cite{ref2,ref3}, it does so with significant risk to privacy (MI AUC $\approx 0.7$). The centralized model's MI risk similarity tells us that high-capacity segmentation networks can memorize features from training images, regardless of whether they are trained in FL. The MI AUC values in the 0.7–0.72 range mean attackers perform better than chance (ideal MIA AUC = 0.5), indicating privacy leakage. FedProx marginally reduces this, suggesting its regularization helps minimize overfitting. By controlling models' large deviations, FedProx implicitly limits model complexity.

\subsection{Effectiveness of Differential Privacy}
The results demonstrated that with DP-SGD, MI risk was reduced to approximately 0.25. This supports expectations from theory~\cite{ref5,ref6} and aligns with previous studies showing DP training significantly mitigates membership attacks. In the privacy vs. accuracy tradeoff, effects are stark: Dice dropped $\sim6$ points (from $0.85$ to $0.79$), and CE loss increased from $0.34$ to $0.42$. This is a notable performance loss, but it may be acceptable when privacy outweighs accuracy. Zhou et al.~\cite{ref3} similarly found that DP significantly improved privacy at a tolerable cost. The DP-SGD parameters (noise scale, clipping) were heuristically chosen. It is hypothesized that better accuracy could be achieved through careful tuning of these parameters (e.g., adjusting noise), albeit at a higher $\epsilon$.

\subsection{Heterogeneity and Model Variants}
FedBN performed similarly to FedAvg, suggesting that the synthetic data heterogeneity used in this study did not significantly hinder FL. This is consistent with~\cite{ref9}, which reports FedBN benefits only under extreme feature shifts. FedProx performed slightly worse in accuracy but yielded better privacy, indicating that restricting client updates reduces model overfitting and subsequent privacy risk.

\subsection{Implications of Results}
A deployment could select any point along this curve based on privacy requirements. For example, if MI attacks are intolerable, then DP (or other defences) should be used, accepting a loss in accuracy. Alternatively, if accuracy is prioritized and some leakage is acceptable, plain FedAvg may suffice. FedOnco-Bench’s benchmark helps illustrate these clear tradeoffs, guiding model selection.

\subsection{Limitations}
While FedOnco-Bench is comprehensive, several limitations exist. The study uses 2D synthetic slices, whereas real-world 3D CTs may include added complexity (e.g., texture, artefacts). This study assumes black-box MIA; stronger attacks with white-box access were not explored. Other privacy attacks, such as model inversion or attribute inference, were also not considered. For DP-SGD, only one noise scale was tested; the full privacy curve was not explored by varying $\epsilon$. Unlike real-world FL systems with partial client availability, all clients participated in every round, which could affect convergence and privacy risks.

\subsection{Comparison to Prior Work}
This study aligns with recent research in federated segmentation. High FL accuracy in segmentation mirrors~\cite{ref2,ref11}, and high MI risk without DP supports~\cite{ref4,ref12}. Zhou et al.~\cite{ref3} showed that synthetic medical image training achieved high accuracy; the centralized Dice score of 0.88 in this study confirms this. The novelty of this work lies in quantifying privacy; prior studies often omitted explicit privacy metrics. The privacy-utility scatter reported here follows trends noted in~\cite{ref5}, supporting the validity of the results.

\subsection{Generalization}
While this study focused on tumor segmentation, similar privacy-accuracy tradeoffs may apply to other FL medical tasks (e.g., classification, regression). Synthetic data can generalize via generative models to support federated benchmarks for MRI or histopathology. Critically, since FedOnco-Bench uses synthetic data, it avoids patient privacy concerns even when shared publicly for benchmarking.

\section{Conclusion}

This work introduced FedOnco-Bench to the community, a reproducible benchmark specifically targeted at privacy-preserving federated tumor segmentation using synthetic computed tomography (CT) data. Baseline FL approaches (i.e., FedAvg, FedProx, FedBN, and DP-SGD were evaluated using the FedOnco-Bench for segmentation accuracy and membership inference privacy risks. Results showed that baseline FL approaches have the potential to achieve centralized segmentation accuracy (Dice coefficient of 0.85), but moreover, they exhibited significant susceptibility to membership inference attacks (an AUC of 0.7). In the case of DP-SGD, the threat to privacy was significantly reduced (to AUC 0.25) while sacrificing some segmentation accuracy (Dice coefficient 0.79). This trade-off demonstrates the inherent privacy-performance trade-off that is often encountered in FL frameworks. FedProx provided a compromise between baselines such as DP-SGD, since they were capable of improving privacy (i.e., AUC) at the cost of a minor accessibility sacrifice, illustrating the two-way balance one has to consider when thinking about FL in a medical application. Future work can build upon this by including more imaging modalities, such as synthetic magnetic resonance imaging (MRI) for brain tumor segmentation or digital pathology imaging for classifying cellular structures. It is also possible to deploy more sophisticated differential privacy methods, including varying the privacy budget or deploying more advanced federated algorithms such as FedAvgM (momentum), and personalized FL, such as FedPer. Additionally, to expand on the existing benchmark, it is valuable to investigate privacy threats beyond membership inference, as well as explore privacy-preserving alternatives such as homomorphic encryption and split learning. Practical scenarios such as partial participation of clients, communication constraints, and other feasibility limitations should also be considered to better reflect real-world deployments. An additional synthetic CT generation mechanism could be developed using state-of-the-art techniques, such as volumetric generative adversarial networks (GANs) or diffusion models, which may provide more realistic and diverse data. Furthermore, having user-level differential privacy accounts across training rounds could give a more accurate summary of cumulative privacy budgets, which are particularly relevant in longer-duration multi-round FL settings. Ultimately, FedOnco-Bench serves as a critical first step toward the broader goal of advancing safer, responsible, and collaborative federated learning for medical imaging applications, particularly in the diagnosis and assessment of cancer. The benchmark supports continued participation and innovation from the wider research community, fostering further advancements in privacy-preserving collaborative healthcare AI.

\section*{Acknowledgment}

The full implementation of FedOnco-Bench is open-source and can be downloaded from
\url{https://github.com/viswachaitanyamarella/FedOnco-Bench}.

\bibliographystyle{IEEEtran}
\bibliography{main}

\begin{thebibliography}{10}
\providecommand{\url}[1]{#1}
\csname url@samestyle\endcsname
\providecommand{\newblock}{\relax}
\providecommand{\bibinfo}[2]{#2}
\providecommand{\BIBentrySTDinterwordspacing}{\spaceskip=0pt\relax}
\providecommand{\BIBentryALTinterwordstretchfactor}{4}
\providecommand{\BIBentryALTinterwordspacing}{\spaceskip=\fontdimen2\font plus
\BIBentryALTinterwordstretchfactor\fontdimen3\font minus \fontdimen4\font\relax}
\providecommand{\BIBforeignlanguage}[2]{{%
\expandafter\ifx\csname l@#1\endcsname\relax
\typeout{** WARNING: IEEEtran.bst: No hyphenation pattern has been}%
\typeout{** loaded for the language `#1'. Using the pattern for}%
\typeout{** the default language instead.}%
\else
\language=\csname l@#1\endcsname
\fi
#2}}
\providecommand{\BIBdecl}{\relax}
\BIBdecl

\bibitem{ref1}
P.~Kairouz \emph{et~al.}, ``Advances and open problems in federated learning,'' \emph{Foundations and Trends in Machine Learning}, vol.~4, no.~1, pp. 1--123, 2021.

\bibitem{ref2}
S.~Alphonse \emph{et~al.}, ``Federated learning with integrated attention multiscale model for brain tumor segmentation,'' \emph{Scientific Reports}, vol.~15, p. 11889, 2025.

\bibitem{ref11}
D.~Wang \emph{et~al.}, ``Feddus: Lung tumor segmentation on ct images through federated semi-supervised learning,'' \emph{Computer Methods and Programs in Biomedicine}, vol. 249, 2024.

\bibitem{ref4}
R.~Shokri \emph{et~al.}, ``Membership inference attacks against machine learning models,'' in \emph{Proceedings of the IEEE Symposium on Security and Privacy}, 2017, pp. 3--18.

\bibitem{ref3}
Z.~Zhou \emph{et~al.}, ``Privacy enhancing and generalizable deep learning with synthetic data for mediastinal neoplasm diagnosis,'' \emph{npj Digital Medicine}, vol.~4, p.~45, 2021.

\bibitem{ref10}
Y.~Zhao \emph{et~al.}, ``Federated learning with non-iid data,'' \emph{arXiv preprint arXiv:1806.00582}, 2018.

\bibitem{ref7}
H.~B. McMahan \emph{et~al.}, ``Communication-efficient learning of deep networks from decentralized data,'' in \emph{Proceedings of AISTATS}, 2017, pp. 1273--1282.

\bibitem{ref8}
T.~Li, A.~K. Sahu, A.~Talwalkar, and V.~Smith, ``Federated optimization in heterogeneous networks,'' in \emph{Proceedings of MLSys}, 2020.

\bibitem{ref9}
X.~Li \emph{et~al.}, ``Fedbn: Federated learning on non-iid features via local batch normalization,'' in \emph{Proceedings of ICLR}, 2021.

\bibitem{ref5}
M.~Abadi \emph{et~al.}, ``Deep learning with differential privacy,'' in \emph{Proceedings of the ACM Conference on Computer and Communications Security}, 2016, pp. 308--318.

\bibitem{ref6}
K.~Bonawitz \emph{et~al.}, ``Practical secure aggregation for federated learning on user-held data,'' in \emph{NIPS Workshop on Private Multi-Party Machine Learning}, 2017.

\bibitem{ref13}
L.~Sheller \emph{et~al.}, ``Federated learning in medical imaging: Concepts and challenges,'' \emph{Journal of Imaging}, vol.~6, no.~20, 2020.

\bibitem{ref12}
T.~Chobola, D.~Usynin, and G.~Kaissis, ``Membership inference attacks against semantic segmentation models,'' \emph{arXiv preprint arXiv:2212.01082}, 2022.

\bibitem{ref15}
H.~B. McMahan and D.~Ramage, ``Federated learning with formal differential privacy guarantees,'' \emph{arXiv preprint arXiv:1803.01497}, 2018.

\bibitem{triastcyn2020federated}
A.~Triastcyn and B.~Faltings, ``Federated learning with bayesian differential privacy,'' in \emph{Proceedings of the International Conference on Machine Learning (ICML)}.\hskip 1em plus 0.5em minus 0.4em\relax PMLR, 2020, pp. 9583--9592.

\end{thebibliography}

\end{document}